\documentclass[10pt,twocolumn,letterpaper]{article}

\usepackage[pagenumbers]{cvpr} 

\usepackage[dvipsnames]{xcolor}

\definecolor{cvprblue}{rgb}{0.21,0.49,0.74}
\usepackage[pagebackref,breaklinks,colorlinks,citecolor=cvprblue]{hyperref}
\usepackage{listings}
\usepackage{xcolor}
\usepackage{bbding}
\usepackage{enumitem,amssymb}
\usepackage{multicol}
\usepackage{multirow}
\usepackage{amssymb}
\usepackage{arydshln}

\newlist{todolist}{itemize}{2}
\setlist[todolist]{label=$\square$}
\usepackage{pifont}
\newcommand{\cmark}{\ding{51}}%
\newcommand{\xmark}{\ding{55}}%
\newcommand{\done}{\rlap{$\square$}{\raisebox{2pt}{\large\hspace{1pt}\cmark}}\hspace{-2.5pt}}
\newcommand{\wontfix}{\rlap{$\square$}{\large\hspace{1pt}\xmark}}

\definecolor{CPPLight}  {HTML} {686868}
\definecolor{CPPSteel}  {HTML} {888888}
\definecolor{CPPDark}   {HTML} {262626}
\definecolor{CPPBlue}   {HTML} {4172A3}
\definecolor{CPPGreen}  {HTML} {487818}
\definecolor{CPPBrown}  {HTML} {A07040}
\definecolor{CPPRed}    {HTML} {AD4D3A}
\definecolor{CPPViolet} {HTML} {7040A0}
\definecolor{CPPGray}  {HTML} {B8B8B8}

\def\paperID{***} 
\def\confName{CVPR}
\def\confYear{2024}

\title{$\mathrm{Hyper}\mathcal{Z{\cdot}Z{\cdot}W}$ Operator Connects Slow-Fast Networks for Full Context Interaction}

\author{Harvie Zhang \\ 
HyperEvol AI Lab \\ 
{\tt\small harvie.zhang@hyperevol.com}
}

\begin{document}
\maketitle

\renewcommand{\thefootnote}{\textdagger}


\renewcommand{\thefootnote}{\arabic{footnote}}

\begin{abstract}

The self-attention mechanism utilizes large implicit weight matrices, programmed through dot product-based activations with very few trainable parameters, to enable long sequence modeling. In this paper, we investigate the possibility of discarding residual learning by employing large implicit kernels to achieve full context interaction at each layer of the network. To accomplish it, we introduce coordinate-based implicit MLPs as a slow network to generate hyper-kernels\footnote[1]{The definition of Hyper-Kernel comes from HyperNEAT~\cite{HyperNEAT2009}.} for another fast convolutional network\footnote[2]{The slow network serves as a hyper-kernel generator, while the fast network either interacts with the context or has context-dependent fast weights. Our definitions of slow and fast networks differ from~\cite{fwp-jurgen1992,fwp2021,slow-fast1987}.}. To get context-varying weights for fast dynamic encoding, we propose a $\mathrm{Hyper}\mathcal{Z{\cdot}Z{\cdot}W}$ operator that connects hyper-kernels ($\mathcal{W}$) and hidden activations ($\mathcal{Z}$) through simple elementwise multiplication, followed by convolution of $\mathcal{Z}$ using the context-dependent $\mathcal{W}$. Based on this design, we present a novel Terminator architecture that integrates hyper-kernels of different sizes to produce multi-branch hidden representations for enhancing the feature extraction capability of each layer. Additionally, a bottleneck layer is employed to compress the concatenated channels, allowing only valuable information to propagate to the subsequent layers. Notably, our model incorporates several innovative components and exhibits excellent properties, such as introducing local feedback error for updating the slow network, stable zero-mean features, faster training convergence, and fewer model parameters. Extensive experimental results on pixel-level 1D and 2D image classification benchmarks demonstrate the superior performance of our architecture. Our code will be publicly available at \href{https://github.com/hyperevolnet/Terminator}{https://github.com/hyperevolnet/Terminator}.

\end{abstract}

\section{Introduction}

\begin{figure}[t]
\centering
    \begin{minipage}{1.0\linewidth}\centering
	   \includegraphics[scale=0.66]{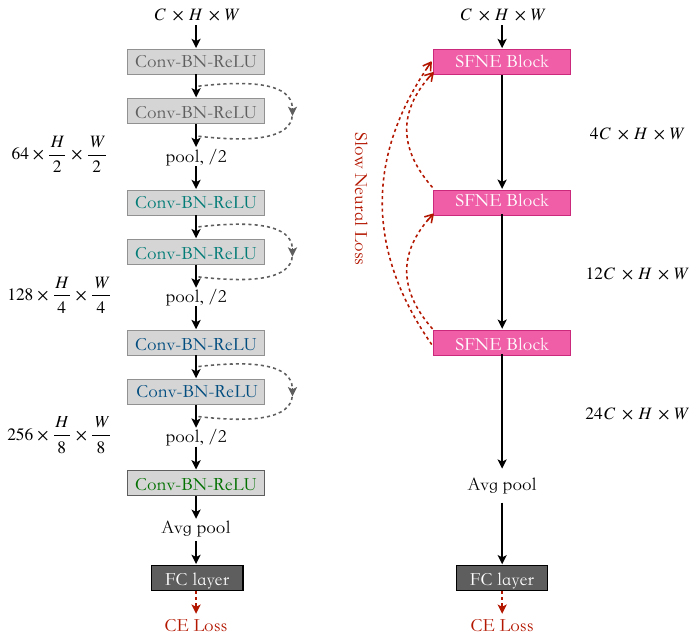} 
    \end{minipage} 
\vspace{-5pt}
\caption{Comparison between the residual network and our Terminator architecture. 1) Our Slow-Fast Neural Encoding (SFNE) block employs a multi-branch structure, eliminating the need for residual learning (Figure~\ref{fig:sfne}). 2) Our hidden layers do not utilize pooling layers for downsampling feature resolution. 3) We introduce a novel local feedback error for updating the slow network.}
\label{fig:bottleneck}
\vspace{-8pt}
\end{figure}

In the past decade, significant advancements have been made in neural networks, particularly with the emergence of ResNet~\cite{resnet2016} and Transformer~\cite{attention2017}. However, the additive connections between layers in residual learning, while addressing the performance degeneration, pose challenges for the exploration of interpretable architectures~\cite{ex-residual-ensemble2016,ex-challenge2022} and the adoption of greedy layerwise training~\cite{local2019}. Mainstream residual networks~\cite{resnet2016,densenet2017,wide-resnet2016,efficientnet2019,convnext2022} typically rely on single-branch architectures with small convolution kernels, which significantly diminish the feature extraction capabilities of individual layers. To compensate information loss, previous layer outputs are added to subsequent layers. However, increasing the size of convolution kernels leads to an exponential growth in model parameters. Consequently, it is crucial to address the following questions in order to eliminate the reliance on residual learning:

\textit{How can we generate large convolution kernels with fewer parameters and build a multi-branch network?}

In the Transformer models~\cite{attention2017,ViT2020,swin2021}, the self-attention weights can be viewed as context-dependent weight matrices or neural network-programmed fast weights~\cite{fwp-jurgen1992,fwp2021,fwp-Ba2016}. These models employ large implicit weight matrices, derived from key-query activations with few trainable parameters, to encode long sequence inputs. However, the utilization of dot product-based attention matrices introduces quadratic time and space complexity, making it challenging to scale Transformers to inputs with large context sizes. Therefore, this paper aims to identify a simpler and more effective method than dot product-based attention to obtain context-varying weights through network activations.

\textit{How can we obtain context-dependent fast weights through simpler elementwise multiplication?}

This paper not only investigates methods to leverage network activations for generating context-dependent large implicit convolution kernels but also proposes a novel multi-branch architecture to explore the possibility of abandoning residual learning. Specifically, we employ coordinate-based implicit MLPs~\cite{mfn2020} with few model parameters as a slow network to generate hyper-kernels for another fast convolutional network. The weights of slow network is data-independent and remain fixed after training. To make the generated hyper-kernels to be context-dependent, we propose a $\mathrm{Hyper}\mathcal{Z{\cdot}Z{\cdot}W}$ operator, connecting hyper-kernels ($\mathcal{W}$) and hidden activations ($\mathcal{Z}$) through simple elementwise multiplication, a more efficient alternative to the dot product attention~\cite{attention2017}. The fast network utilizes context-varying weights to perform convolution operations on $\mathcal{Z}$. Consequently, the weights of fast network are context-dependent and can change at test time. Based on above design, we present a new architecture called Terminator, which integrates context-dependent hyper-kernels of different sizes to produce multi-branch hidden representations, enhancing the feature extraction capability of each layer. Additionally, a bottleneck layer is employed to compress the concatenated channels, allowing only valuable information to propagate to the subsequent layers. 

Furthermore, several novel components are proposed to build the multi-branch architecture, including Recursive Gated Unit (RGU), hyper-channel interaction and hyper interaction (see Figure~\ref{fig:hyper_interaction}). We also introduce a local feedback loss for updating the slow network and replace the normalization layer that discards \textit{affine} parameters and \textit{momentum} argument with a standardization layer. Our architecture exhibits excellent properties such as stable zero-mean features, faster training convergence, and fewer model parameters. Extensive experimental results on pixel-level 1D and 2D image classification benchmarks demonstrate the superior performance of our architecture.

\textbf{Notation:} Let $\mathbf{x} \in \mathbb{R}^{C \times H \times W}$ be the 2D input, where $H$, $W$ and $C$ are height, width, and channel dimension. The $\mathrm{Hyper}\mathcal{Z{\cdot}Z{\cdot}W}$ can be abbreviated as HyperZZW.

\begin{figure}[t]
\centering
    \begin{minipage}{1.0\linewidth}\centering
        \includegraphics[scale=0.395]{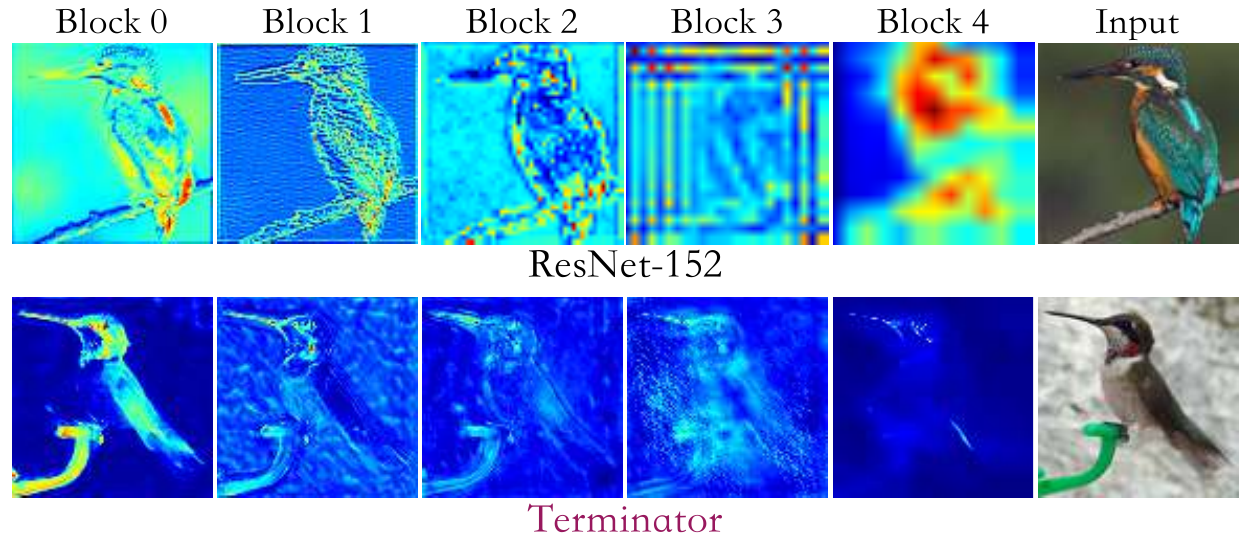}
    \end{minipage} 
\vspace{-7pt}
\caption{Visualization of the feature maps in each block. For the convenience of comparison, we enlarge the output of the 2$\sim$4 blocks of ResNet-152.} 
\label{fig:vis_res_termi}
\vspace{-8pt}
\end{figure}

\section{Method}
\label{sec:method}
In this section, we first provide an overview of the key properties of our overall architecture. Subsequently, we present the specifics of the coordinate-based implicit MLPs served as a slow network for generating hyper-kernels, and elucidate the role played by the $\mathrm{Hyper}\mathcal{Z{\cdot}Z{\cdot}W}$ operator. Following that, we elaborate on the process of constructing a multi-branch SFNE block using our proposed novel components. Lastly, we introduce the slow neural loss.

\begin{figure*}[t]
\vspace{-12pt}
\centering
    \begin{minipage}{1.0\linewidth}\centering
	   \includegraphics[scale=0.79]{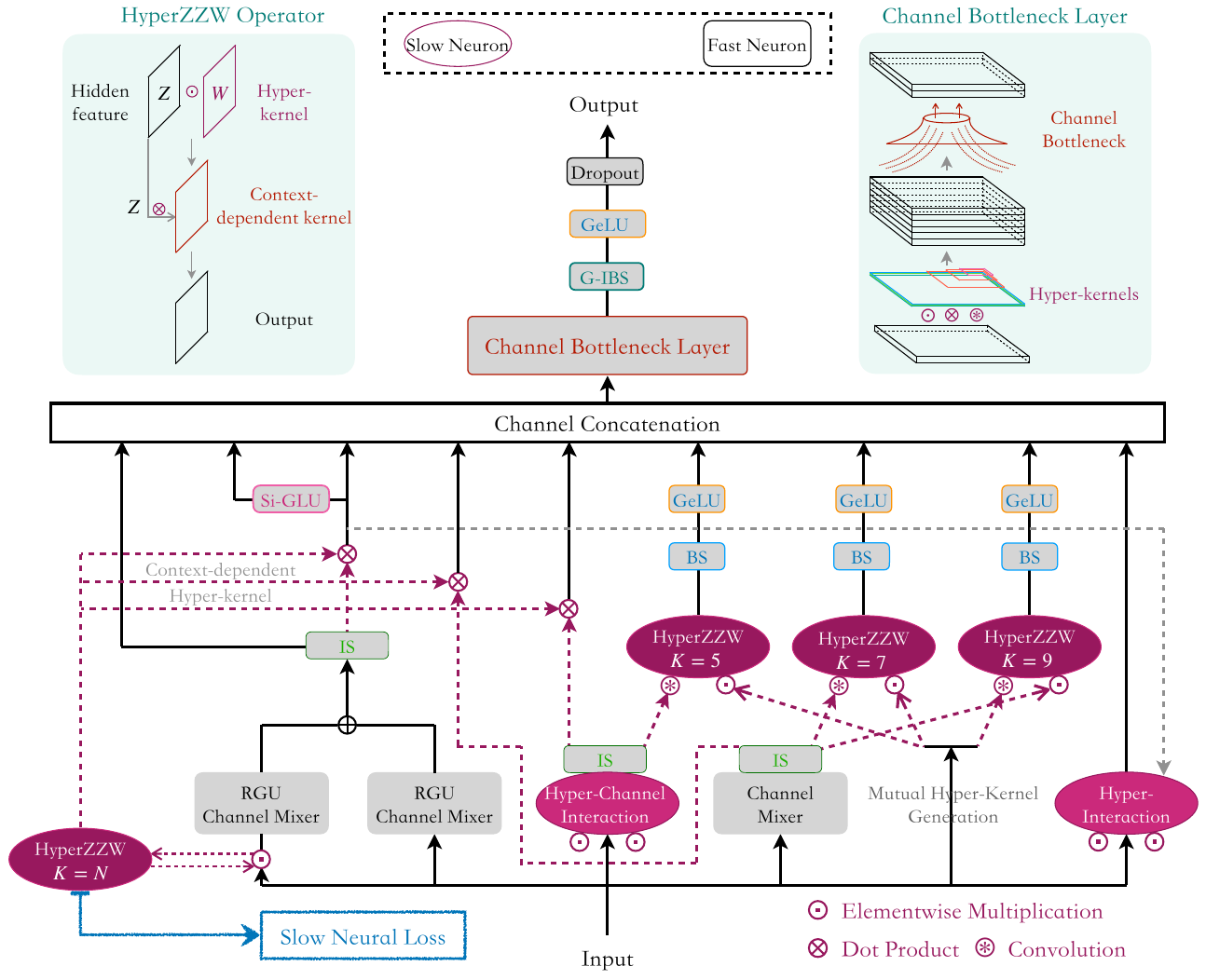}
    \end{minipage} 
\vspace{-5pt}
\caption{The overall framework of our Slow-Fast Neural Encoding (SFNE) block, which utilizes channel mixers and multi-scale hyper-kernels (e.g. $N$ is the input size) to construct a nine-branch structure. The ovals represent slow networks used to generate hyper-kernels, and the rectangles represent fast networks that interact directly with the input. The $\mathrm{Hyper}\mathcal{Z{\cdot}Z{\cdot}W}$ operator is formed by combining $\odot$, $\otimes$, and $\circledast$, enabling context-dependent fast weights. RGU and Si-GLU represent the recursive gated unit and the simplified gated linear unit.}
\label{fig:sfne}
\vspace{-5pt}
\end{figure*}

\subsection{Key Properties}

\begin{figure}
\centering
    \begin{minipage}{1.0\linewidth}\centering
        \includegraphics[scale=0.45]{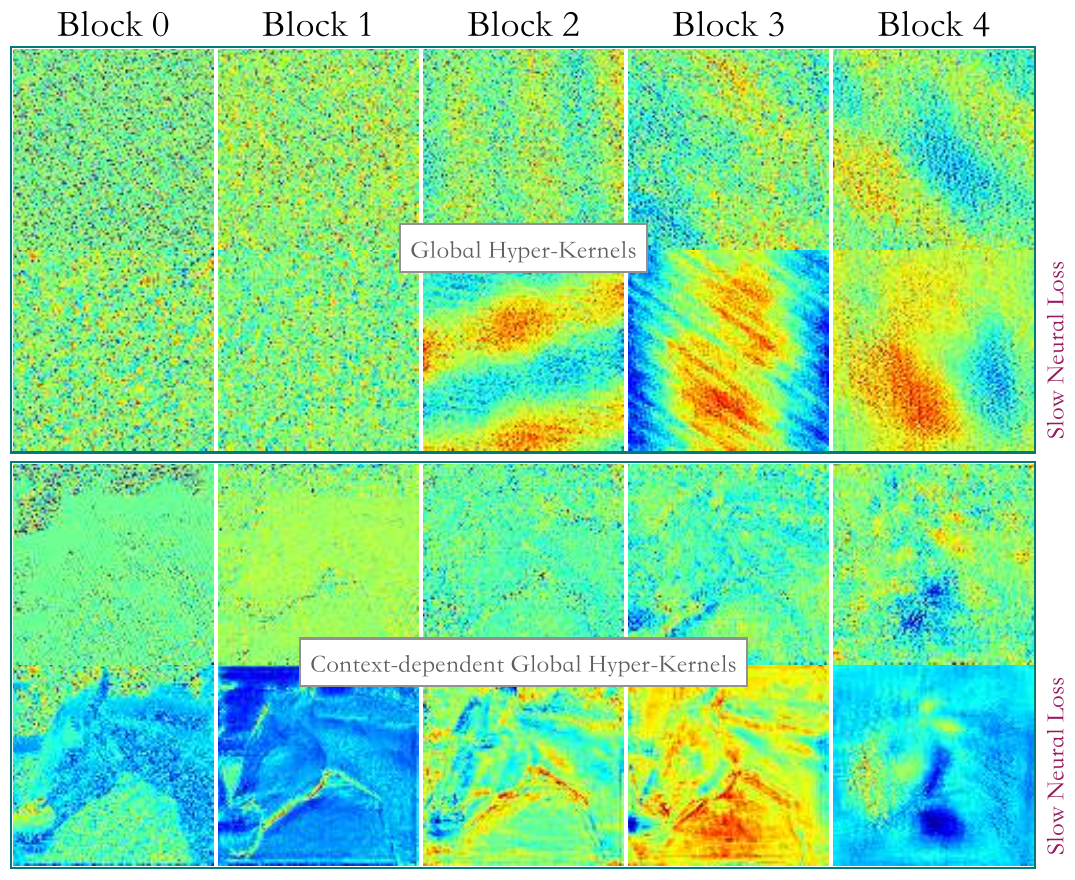}
    \end{minipage} 
\vspace{-5pt}
\caption{Visualization of global hyper-kernels in each block. By performing elementwise multiplication between the sample activations and the global hyper-kernels $\mathbf{K}_g$, the model can effectively leverage context-dependent hyper-kernels $\hat{\mathbf{K}}_g$ to obtain pixel-level scores, especially when trained with the slow neural loss.}
\label{fig:vis_no_slow_loss}
\vspace{-8pt}
\end{figure}

\begin{itemize}
    \item[\wontfix] \textbf{[No Residual Learning.]} Residual learning has been widely employed to address the performance degeneration problem in very deep neural networks~\cite{resnet2016,wide-resnet2016,resnext2017,efficientnet2019,convnext2022}. In this paper, we content that the crucial motivation behind the adoption of residual connections is to compensate for the limited representation learning capacity of individual layers, particularly in the case of residual networks with its single-branch architecture and small convolution kernels. To overcome these limitations, we incorporate large implicit convolution kernels of varying sizes to construct a multi-branch network, enabling full context interaction at each layer and obviating the necessity for residual learning. Based on above analysis, we introduce a novel Terminator architecture, which is composed of a stack of SFNE blocks shown in Figure~\ref{fig:sfne}.

    \item[\wontfix] \textbf{[No Dot Product Attention.]} The quadratic increase in computational cost, based on the number of pixels, necessitates the division of images into patches in order to facilitate the successful implementation of vision transformers~\cite{ViT2020}. However, the conversion of patch embeddings into fixed-length representations not only limits the inclusion of local context information but also hinders the estimation of attention scores for individual pixels. In contrast, our proposed $\mathrm{Hyper}\mathcal{Z{\cdot}Z{\cdot}W}$ operator enables the interaction between generated hyper-kernels and image pixels through a simple elementwise multiplication, facilitating the acquisition of precise pixel-level (i.e. token-level) scores (see Figure~\ref{fig:vis_no_slow_loss}). Notably, this approach is more efficient than the attention-free transformer~\cite{attention-free2021} as it eliminates the reliance on the Sigmoid function.

    \item[\wontfix] \textbf{[No Intermediate Pooling Layer.]} Intermediate pooling layers enhance the receptive field by reducing feature map size, enabling better capture of global information. However, our visualizations in Figure~\ref{fig:vis_res_termi} demonstrate that the intermediate pooling layers can distort the overall structure of objects.\footnote[3]{To obtain 2D heat map, we sum the values of all channels in the feature map. The visualization in Figure \ref{fig:vis_no_slow_loss} follows the same process.} Therefore, this hampers final average pooling layer to generate accurate image descriptors. In our architecture, the utilization of large implicit convolutional kernels enables global contextual interaction at each layer, obviating the need for pooling layers to expand the receptive field. Moreover, the feature maps produced by our architecture exhibit a notable continuous denoising process (see Figure~\ref{fig:vis_res_termi}), which not only alleviates the model's learning complexity but also facilitates the development of interpretable neural networks.

    \item[\wontfix] \textbf{[No Normalization.]} Traditionally, normalization layers such as BN~\cite{BN2015}, IN~\cite{IN2016}, LN~\cite{LN2016}, GN~\cite{GN2018} incorporate learnable \textit{affine} parameters, including scaling factor $\gamma$ and shift factor $\beta$. Previous work~\cite{BN2015} asserted that these parameters restore the network's representation power. However, our experimental findings, as illustrated in Figure~\ref{fig:stl-acc_bn_bs}, reveal that their primary function is to correct batch statistics (mean and variance) to ensure stable activations. BN~\cite{BN2015} and IN~\cite{IN2016} also employ a \textit{momentum} argument to reduce volatility between batches, but it can hinder the in-context learning in each batch. 
    In this paper, we remove the \textit{affine} and \textit{momentum} parameters, transforming normalization into a standardized operation called z-score standardization. The new standardization can be written as:
    \begin{equation}
        \hat{x} = \frac{x - \mathrm{E}[x]}{\sqrt{Var[x]  + \epsilon}},
    \end{equation}
    where $\epsilon$ is a constant added to the mini-batch variance for numerical stability. Consequently, we replace the original BN and IN layers with Batch Standardization (BS) and Instance Standardization (IS) respectively.

    \item[\done] \textbf{[Slow-Fast Networks.]} In our architecture, we employ coordinate-based implicit MLPs~\cite{mfn2020} as a slow network to generate large convolution kernels (hyper-kernels) for a fast network. The slow network has minimal trainable parameters (see Table~\ref{tab:model_params}) but can generate large hyper-kernels to achieve full context interaction at each layer. Notably, the weights of the slow network remain fixed after training and do not directly interact with the context~\cite{fwp2021}. In contrast, as depicted in Figure~\ref{fig:sfne}, the fast network consists of two components. The first component is the channel mixers and bottleneck layer, which directly interact with the feature maps. The second component involves the context-dependent fast weights generated by the $\mathrm{Hyper}\mathcal{Z{\cdot}Z{\cdot}W}$ operator, enabling convolution and dot product operations. These fast weights can adapt in real-time during testing based on the input.
    
\end{itemize}

\subsection{Generation of Hyper-kernels}
To address the first question presented in the introduction section, we generate two distinct hyper-kernels using slow networks to construct multi-branch networks. The first one, denoted as global hyper-kernel $\mathbf{K}_g \in \mathbb{R}^{1 \times C \times H \times W}$, has the same size as the input $\mathbf{x}$. The second one, referred to as local hyper-kernel is $\mathbf{K}_l \in \mathbb{R}^{C \times 1 \times k \times k}$ for depth-wise convolution, has small kernel size $k$. By combining these hyper-kernels, our architecture can extract global features while preserving the input's fine details. Below we elaborate on the generation process of $\mathbf{K}_g$.

The global hyper-kernel $\mathbf{K}_g \in \mathbb{R}^{1 \times C \times H \times W}$ is generated using coordinate-based implicit MLPs, which is also known as multiplicative filter networks~\cite{mfn2020}. The generation process is described by the following equations:
\begin{align}
    \mathbf{h}^{(1)} &= \mathit{g} \left(\mathbf{c}; \mathbf{\theta}^{(1)}\right) \nonumber \\
    \mathbf{h}^{(i+1)} &= \left(\mathbf{W}^{(i)}\mathbf{h}^{(i)} + b^{(i)} \right) \odot \mathit{g} \left(\mathbf{c}; \mathbf{\theta}^{(i+1)}\right), i=1,...,l \nonumber \\
    \mathbf{h}^{(o)} &= \mathbf{W}^{(o)}\mathbf{h}^{(i+1)} + b^{(o)},  \label{eq:mpf}
\end{align}
where $\odot$ denotes elementwise multiplication, $\mathbf{c} \in \mathbb{R}^{1 \times 2 \times H \times W}$ represents the normalized coordinates on the $H$ and $W$ dimensions, which lie in a uniform linear space of $[-1, 1]^H$ and $[-1, 1]^W$. The weight matrix for the $i$-th layer is denoted by $\mathbf{W}^{(i)} \in \mathbb{R}^{d \times d}$, and the hidden unit of the $i$-th layer is $\mathbf{h}^i \in \mathbb{R}^{1 \times d \times H \times W}$. The weight $W^{(o)} \in \mathbb{R}^{C \times d}$ in the output layer transforms $\mathbf{h}^{i+1}$ to $\mathbf{h}^o \in \mathbb{R}^{1 \times C \times H \times W}$. The function $\mathit{g} \left(\mathbf{c}; \mathbf{\theta}^{(i)}\right)$ is a simple sinusoidal filter defined as:
\begin{equation}
    \mathit{g} \left(\mathbf{c}; \mathbf{\theta}^{(i)}\right) = \text{sin}(\mathbf{\mathit{w}}^{(i)} \mathbf{c} + \mathbf{\phi}^{(i)}),
\end{equation}
where $\theta^{(i)} = {\mathbf{\mathit{w}}^{(i)} \in \mathbb{R}^{d \times 2}, \mathbf{\phi}^{(i)} \in \mathbb{R}^{d}}$.

In addition, we use a simple \textit{s-renormalization}\footnote[4]{The \textit{s-renormalization} is called Weight Normalization in~\cite{weight-norm2016}, and we use it to renormalize the hidden output of the slow network.} operation on the hidden output $\mathbf{h}^{(i+1)}$ to stabilize training of the slow network. This operation is defined as follows:
\begin{equation}
    \hat{\mathbf{h}}^{(i+1)} = \text{diag}(\mathbf{a})\mathbf{I} + \text{diag}(\mathbf{b})\mathbf{h}^{(i+1)}_{\text{norm}},
\end{equation}
where $\mathbf{a} \in \mathbb{R}^{d}$ and $\mathbf{b} \in \mathbb{R}^{d}$ are scaling vectors learned during training. $\mathbf{h}^{(i+1)}_{\text{norm}}$ denotes the Euclidean norm of $\mathbf{h}^{(i+1)}$. We initialize $\mathbf{a}$ and $\mathbf{b}$ to 1.0 and 0.1, respectively, following the setup described in~\cite{diracnets2017}.

\subsection{HyperZZW Operator}
Our proposed $\mathrm{Hyper}\mathcal{Z{\cdot}Z{\cdot}W}$ operator enables the generated global and local hyper-kernels to interact with the context, resulting in context-dependent fast weights. The utilization of $\mathrm{Hyper}\mathcal{Z{\cdot}Z{\cdot}W}$ involves two steps. First, hidden activation $\mathcal{Z}$ and hyper-kernel $\mathcal{W}$ are multiplied elementwise to obtain context-varying weights $\hat{\mathcal{W}}$. Subsequently, $\hat{\mathcal{W}}$ can be used to perform dot product or convolution operations on $\mathcal{Z}$ for feature extraction. Unlike dot product-based attention~\cite{attention2017}, our $\mathrm{Hyper}\mathcal{Z{\cdot}Z{\cdot}W}$ operator achieves context-dependent fast weights with linear time and space complexity. The following steps outline the specific usage $\mathrm{Hyper}\mathcal{Z{\cdot}Z{\cdot}W}$ on hyper-kernels $\mathbf{K}_g$ and $\mathbf{K}_l$.

\textbf{[Global HyperZZW.]} For the global hyper-kernel $\mathbf{K}_g$, we employ a simple elementwise multiplication to get context-dependent $\hat{\mathbf{K}}_g$:
\begin{equation}
    \hat{\mathbf{K}}_g = \mathbf{Z} \odot \mathbf{K}_g,
\end{equation}
where $\mathbf{Z} \in \mathbb{R}^{B \times C \times H \times W}$ represents $B$ samples in a batch, and $\hat{\mathbf{K}}_g \in \mathbb{R}^{B \times C \times H \times W}$ due to broadcasting $\mathbf{K}_g$ along the batch dimension. The visualization results, shown in Figure~\ref{fig:vis_no_slow_loss}, demonstrate the effectiveness of $\hat{\mathbf{K}}_g$ in capturing pixel-level weighting scores for each batch sample, especially when the model is trained with slow neural loss (as described in Section~\ref{sec:slow_neural_loss}). In contrast, the context-independent $\mathbf{K}_g$ generated by the slow neural network appears to resemble random noise. Subsequently, instead of using sliding window-based convolution, we employ the dot product $\otimes$ operation with $\hat{\mathbf{K}}_g$ for global feature extraction:
\begin{equation}
    \mathbf{Z}_g = \mathbf{Z} \otimes \hat{\mathbf{K}}_g,
    \label{eq:dot}
\end{equation}
where $\mathbf{Z}_g \in \mathbb{R}^{B \times C \times H \times W}$. For 1D samples $\mathbf{Z} \in \mathbb{R}^{B \times C \times L}$ (e.g. $L$ is the sequence length), we can utilize the following Fast Fourier Transform (FFT) to extract global features:
\begin{equation}
    \mathbf{Z}_g = \text{iFFT}(\text{FFT}(\mathbf{Z}) \odot \text{FFT}(\hat{\mathbf{K}}_g)),
    \label{eq:fft}
\vspace{-3pt}
\end{equation}

\textbf{[Local HyperZZW.]} To introduce context-dependency to the local hyper-kernel $\mathbf{K}_l \in \mathbb{R}^{C \times 1 \times k \times k}$, we incorporate the $\mathrm{Hyper}\mathcal{Z{\cdot}Z{\cdot}W}$ operator into its generation process. The process is described by the following equations:
\begin{align}
    \hat{\mathbf{Z}}^{(1)} &= \mathit{T} \left(\mathbf{Z}\right) \nonumber \\
    \mathbf{h}^{(1)} &= \mathit{g} \left(\mathbf{c}; \mathbf{\theta}^{(1)}\right) \nonumber \\
    \mathbf{h}^{(i+1)} &= \left(\mathbf{W}^{(i)}\mathbf{h}^{(i)} + b^{(i)} \right) \odot \mathit{g} \left(\mathbf{c}; \mathbf{\theta}^{(i+1)}\right), i=1,...,l \nonumber \\ 
    \mathbf{h}^{(i+1)} &= \left(\mathbf{W}_z^{(i)}\hat{\mathbf{Z}}^{(i)} + b_z^{(i)} \right) \odot \mathbf{h}^{(i+1)} \label{eq:mpf-local} \\
    \mathbf{h}^{(o)} &= \mathbf{W}^{(o)}\mathbf{h}^{(l)} + b^{(o)} \nonumber
\vspace{-3pt}
\end{align}
where $\mathbf{c} \in \mathbb{R}^{1 \times 2 \times k \times k}$ represents the normalized coordinates on the $k$ dimension, which range from $-1$ to $1$ in a uniform linear space. The hidden unit of the $i$-th layer is $\mathbf{h}^i \in \mathbb{R}^{1 \times d \times k \times k}$, and the weight $W^{(o)} \in \mathbb{R}^{C \times d}$ in the output layer transforms $\mathbf{h}^{i+1}$ to $\mathbf{h}^o \in \mathbb{R}^{1 \times C \times k \times k}$. The transformation $\mathit{T}$ converts $\mathbf{Z} \in \mathbb{R}^{B \times C \times H \times W}$ to $\hat{\mathbf{Z}} \in \mathbb{R}^{1 \times d \times k \times k}$ through average pooling ($B \times C \times k \times k$), channel reduction ($B \times d \times k \times k$) and batch mean ($1 \times d \times k \times k$). By performing elementwise multiplication between $\hat{\mathbf{Z}}^{(i)}$ and $\mathbf{h}^{(i+1)}$, the context-dependent local hyper-kernels are generated. We can then utilize $\hat{\mathbf{K}}_l \in \mathbb{R}^{C \times 1 \times k \times k}$ to extract local features through sliding window-based convolution operations:
\begin{equation}
    \mathbf{Z}_l = \mathbf{Z} \circledast \hat{\mathbf{K}}_l,
    \label{eq:dot}
\vspace{-3pt}
\end{equation}
where $\circledast$ represents depth-wise convolution. 

Our visualizations in the appendix show that the global features $\mathbf{Z}_g$ primarily capture the overall outline of object, providing a rough representation. On the other hand, local features $\mathbf{Z}_l$ excel at preserving finer details, particularly when smaller convolution kernels are utilized.
To address the challenge of information loss at each layer, the combination between global and local features through channel stacking is used to enhance the comprehensive representation of the network. As a result, the necessity for residual connections to compensate for this loss becomes obsolete.

\setcounter{footnote}{4}

\subsection{Slow-Fast Neural Encoding Block}
Our SFNE block, as illustrated in Figure~\ref{fig:sfne}, plays a crucial role in simultaneously achieving full context interaction in channel and spatial dimensions, so that the network does not lose information at each layer.
The Figure~\ref{fig:sfne} presents a nine-branch architecture, composed of three branches from global $\mathrm{Hyper}\mathcal{Z{\cdot}Z{\cdot}W}$, three branches from local $\mathrm{Hyper}\mathcal{Z{\cdot}Z{\cdot}W}$, one branch from the Si-GLU~\footnote{The parameters-free Si-GLU can be written as $Si$-$GLU(\mathbf{x}) = \sigma(\mathbf{x}) \odot \mathbf{x}$, where $\sigma$ denotes the Sigmoid function.}, one branch from the middle layer, and the final one from hyper interaction. Notably, the global branches share a common hyper-kernel. Moreover, the nine-branch architecture is flexible. To handle higher resolution images, we can adjust the number and kernel size of the local $\mathrm{Hyper}\mathcal{Z{\cdot}Z{\cdot}W}$ while keeping the other modules unchanged. To facilitate communication across channels, as the dot product and convolution operations in $\mathrm{Hyper}\mathcal{Z{\cdot}Z{\cdot}W}$ are channel-wise, we introduce three types of channel transformations: 1) Channel Mixer is a straightforward one-layer MLP; 2) Recursive Gated Unit (RGU); 3) Hyper-Channel Interaction. Following these transformations, instance standardization (Mixer-IS) is applied to ensure the independence of the channel dimension and the sample dimension. In the subsequent sections, we provide a detailed introduction to our proposed novel components.


\textbf{Recursive Gated Unit} is an extension of the gated linear unit (GLU)~\cite{glu2017} that incorporates recursion. 
\begin{equation}
\begin{split}
    \mathbf{K} &= \mathbf{W}^K \mathbf{x}, \\
    \mathbf{V} &= \mathbf{W}^V \mathbf{x}, \\
    \mathbf{Q} &=  \sigma \left(  \textit{S} \left( \mathbf{K} \odot \mathbf{W} \mathbf{V} \right) \right) + b, \\
    \mathbf{Y} &= \mathbf{W}^Y \mathbf{Q},
\end{split}
\end{equation}
where $\sigma$ represents the Gaussian Error Linear Unit (GeLU) nonlinearity, and \textit{S} denotes instance standardization. The outputs $\mathbf{K}$, $\mathbf{V}$, $\mathbf{Q}$, and $\mathbf{Y}$ have the same size as the input $\mathbf{x}$. We apply the RGU to the input $\mathbf{x}$ and context-dependent $\hat{\mathbf{K}}_g$ and then use their sum as the output of a channel mixer.

\begin{figure*}[t]
\vspace{-8pt}
\centering
    \begin{minipage}{1.0\linewidth}\centering
	   \includegraphics[scale=0.325]{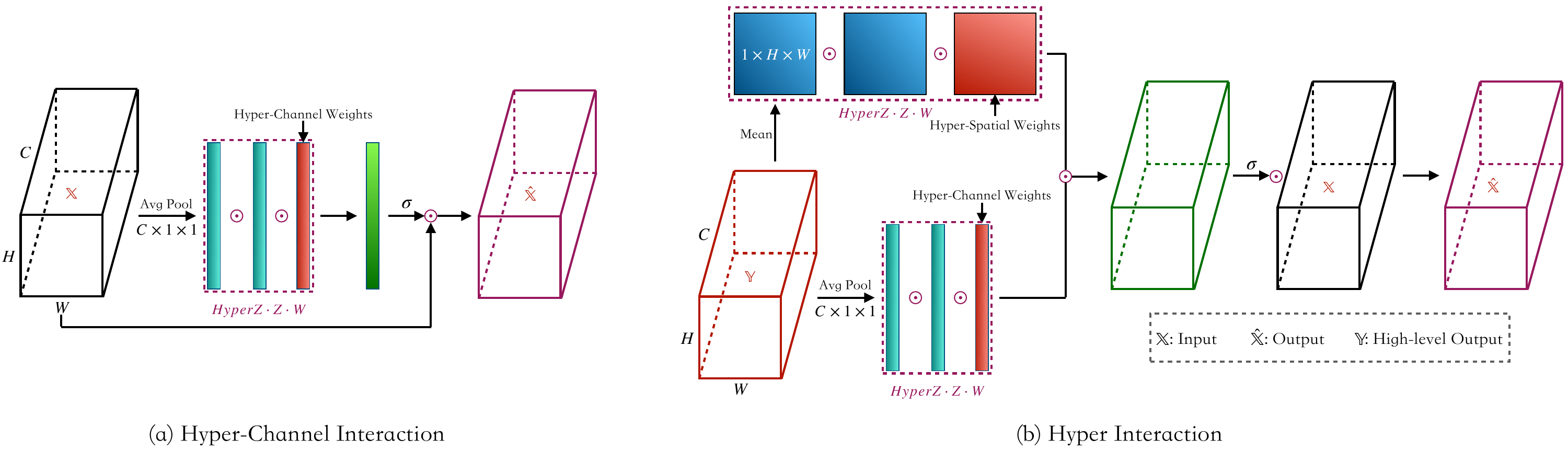}
    \end{minipage} 
\vspace{-8pt}
\caption{Diagrams illustrate our proposed hyper-channel interaction and hyper interaction mechanisms in our architecture.}
\label{fig:hyper_interaction}
\end{figure*}

\begin{table*}
\vspace{-2pt}
  \centering
  \begin{tabular}{ll|ccccc|ll}  
    \toprule
    \multicolumn{2}{c}{Block}    & 0   & 1   & 2   & 3   & 4   & \multicolumn{2}{c}{All}       \\
    
    \hline

    \multicolumn{2}{c|}{Slow Network}     & 73K    & 77K    & 96K     & 153K    & 237K   & 636K  & 8\%  \\

    \cdashline{1-2}[1pt/1pt]

    \multirow{2}*{Fast Network}  
    & Channel mixers & 20K    & 27K    & 92K     & 650K    & 2.7M   & 3.5M  & 45\% \\

    \cdashline{2-2}[1pt/1pt]
    & Bottleneck layer  & 816    & 19K    & 186K    & 1.1M    & 2.4M   & 3.7M  & 47\% \\

    \hline

    \multicolumn{2}{c|}{\multirow{2}*{All}}
    
    & 93K      & 123K    & 374K     & 1.9M      & 5.3M      &  7.8M  \\

    & & 1.2\%    & 1.6\%   & 4.8\%    & 24.4\%    & 68.0\%    &     &  100\%  \\

    \bottomrule
  \end{tabular}
  \vspace{-3pt}
  \caption{Statistics of model parameters in each block, including slow network and fast network. In order to better know the parameters of each module, we list channel mixers and bottleneck layer separately from the fast network.}
  \label{tab:model_params}
  \vspace{-8pt}
\end{table*}


\textbf{Hyper-Channel Interaction} module integrates $\mathrm{Hyper}\mathcal{Z{\cdot}Z{\cdot}W}$ operator, which can be viewed as a more effective version of channel attention mechanisms~\cite{senet2018,eca2020}. 
As depicted in Figure~\ref{fig:hyper_interaction}(a), we begin by aggregating spatial information from a feature map $\mathbf{x} \in \mathbb{R}^{C \times H \times W}$ using an average-pooling operation to generate a channel descriptor $\mathbf{z}_c \in \mathbb{R}^{C \times 1 \times 1}$. Next, our $\mathrm{Hyper}\mathcal{Z{\cdot}Z{\cdot}W}$ operator connects the channel-wise hyper-weights $\mathbf{w}_c \in \mathbb{R}^{C \times 1 \times 1}$ and $\mathbf{z}_c$ generated from a slow network through two simple elementwise multiplications, obtaining gating scores for the channels. Finally, the channel weights are obtained by applying a Sigmoid function, and these weights are broadcasted across the spatial dimension for multiplication with $\mathbf{x}$. Through global cross-channel interaction, our hyper-channel interaction module achieves channel mixing.

\textbf{Hyper Interaction} combines hyper-channel and hyper-spatial interactions into a single module, leveraging the gating scores of the high-level output to automatically extract features from input $\mathbf{x}$. Figure~\ref{fig:hyper_interaction}(b) illustrates this process. To extract channel weights, we employ the same mechanism as described in the hyper-channel interaction. Additionally, for generating the spatial descriptor $\mathbf{z}_s \in \mathbb{R}^{1 \times H \times W}$, we aggregate the channel information of a feature map $\mathbf{x} \in \mathbb{R}^{C \times H \times W}$ using a channel mean operation. Our $\mathrm{Hyper}\mathcal{Z{\cdot}Z{\cdot}W}$ operator connects the spatial-wise hyper-weights $\mathbf{w}_s \in \mathbb{R}^{1 \times H \times W}$ with $\mathbf{z}_s$ through two simple elementwise multiplications, resulting in the spatial gating scores. Next, these scores are multiplied with the channel scores, after broadcasting, to pass through the Sigmoid function for obtaining the channel-spatial weights. Finally, these weights calculated from the high-level output are utilized for pixel-level filtering on the input $\mathbf{x}$. This integration enables the hyper-interaction module to enhance the feature extraction capability of the SFNE block as a new branch.


\textbf{Mutual Hyper-Kernel Generation} (MuHKGen) module can increases the model's complexity and helps prevent over-fitting. For instance, the slow network integrates $\mathbf{z}_2$ to produce a local hyper-kernel $\mathbf{K}_l^1$ using Eqn.~\ref{eq:mpf-local}, and vice versa. Subsequently, the fast networks perform convolution operations on $\mathbf{z}_1$ and $\mathbf{z}_2$ using $\mathbf{K}_l^2$ and $\mathbf{K}_l^1$, respectively, to extract multi-scale features.

\textbf{Channel Bottleneck Layer.} The SFNE block concatenate the multi-scale features along the channel dimension to achieve a comprehensive representation of the input. The objective of the bottleneck layer is to compress the concatenated global and local feature maps, effectively eliminating noise present in the feature maps. To accomplish this, we introduce a bottleneck coefficient $\lambda$ to control the degree of channel compression, determining the reduction in the number of concatenated channels. On the other hand, it scales up the channel dimension of the input by a factor of $\lambda$ ($\lambda \geq 1$). Specifically, when the input is $\mathbf{x} \in \mathbb{R}^{C \times H \times W}$, passing through the bottleneck layer results in an output $\mathbf{y} \in \mathbb{R}^{\lambda C \times H \times W}$.

\begin{figure*}
\vspace{-25pt}
\centering
    \begin{minipage}{0.325\linewidth}\centering
	   \includegraphics[scale=0.385]{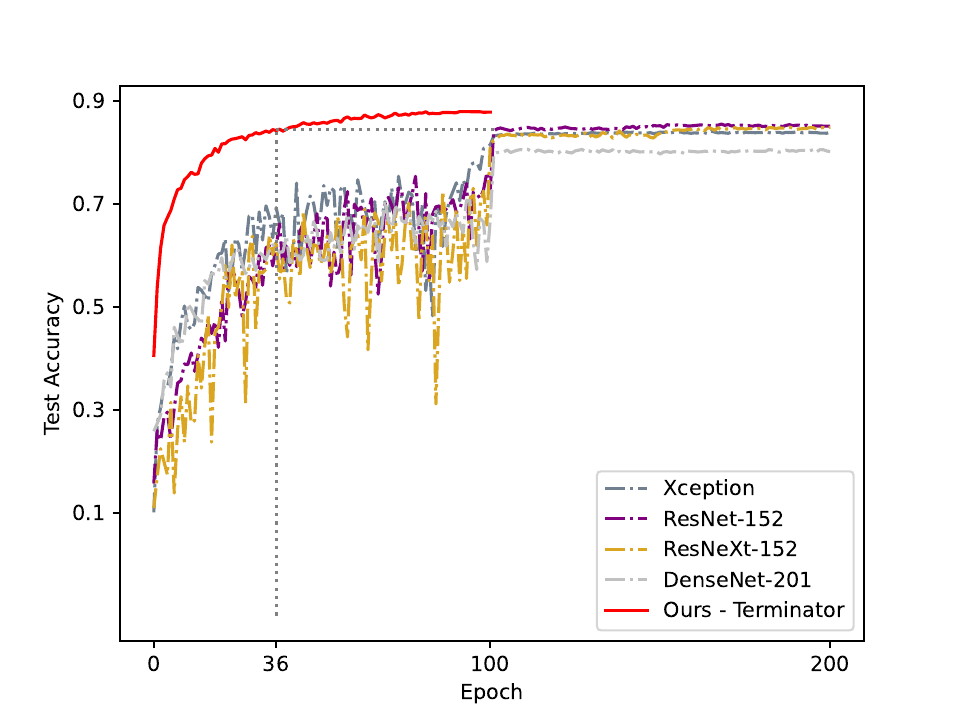} \\
    (a) Training convergence
    \end{minipage} \hspace{1pt}
    \begin{minipage}{0.325\linewidth}\centering
        \includegraphics[scale=0.385]{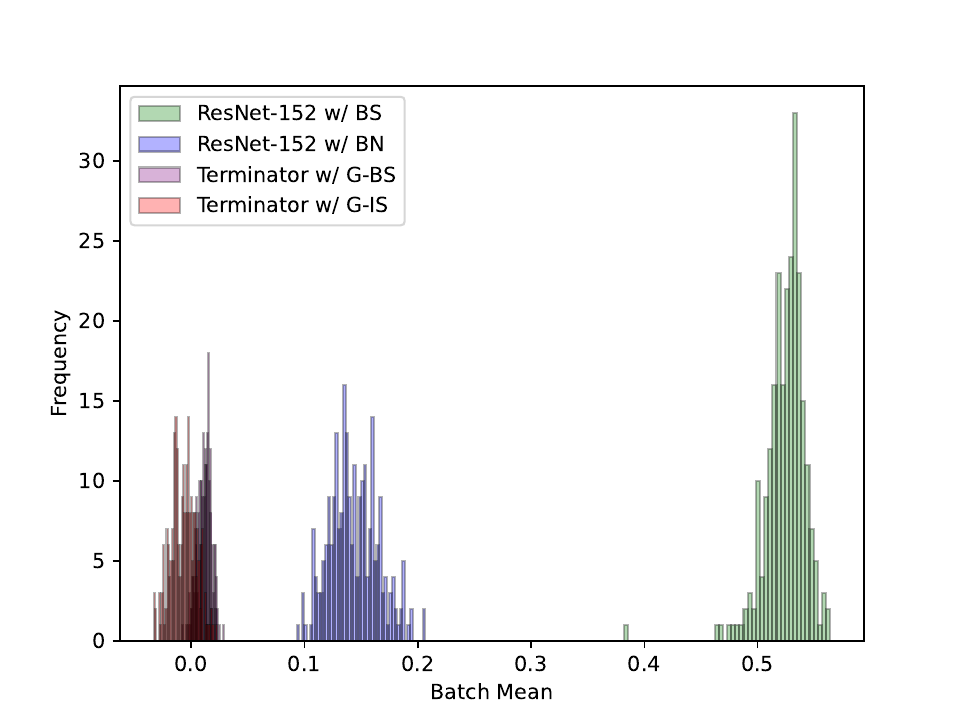} \\
    (b) Channel mean
    \end{minipage} \hspace{1pt}
    \begin{minipage}{0.325\linewidth}\centering
        \includegraphics[scale=0.385]{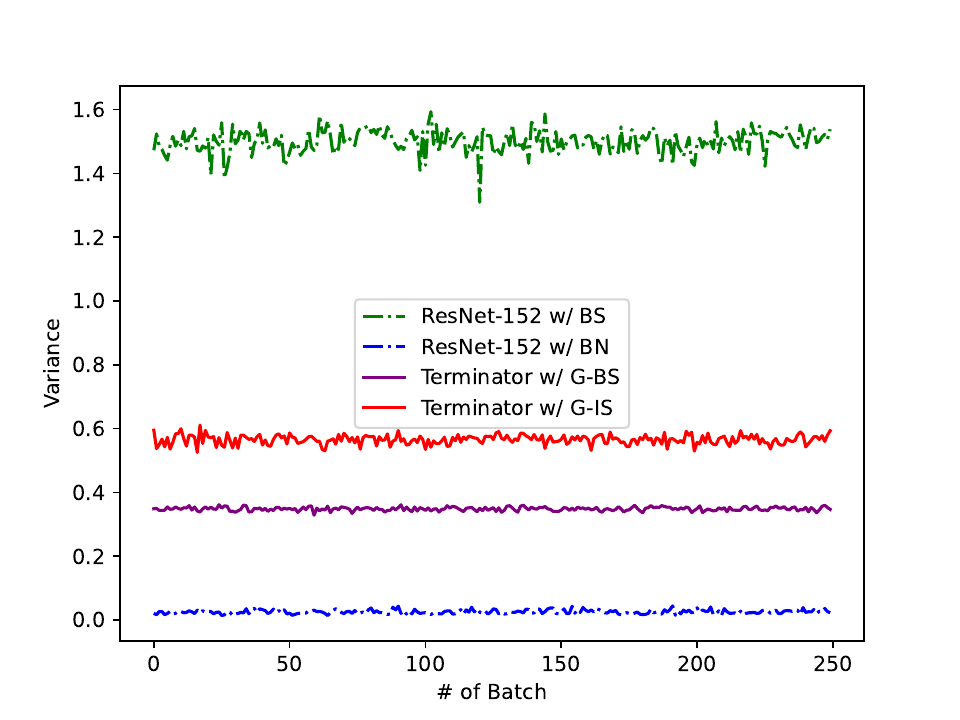} \\
    (c) Channel variance
    \end{minipage} 
\vspace{-4pt}
\caption{(a) shows fast training convergence of our Terminator. (b) and (c) visualize the batch-based channel mean and channel variance statistics of the last hidden layer in ResNet-152 and Terminator. The experiments are constructed on the STL10 test set.}
\label{fig:stl-acc_bn_bs}
\vspace{-3pt}
\end{figure*}

\textbf{Group-based Standardization.} After the channel bottleneck layer, we introduce a Group-based Instance-Batch Standardization (G-IBS) operation, as illustrated in Figure~\ref{fig:sfne}. Specifically, the output feature maps are divided into non-overlapping groups based on channels, and Group-based Batch Standardization (G-BS) and Group-based Instance Standardization (G-IS) are applied alternately to each corresponding group. Our visualized results shown in Figure~\ref{fig:stl-acc_bn_bs}(b-c) demonstrate that combining G-IS and G-BS can enhance the model's expressive power by increasing diversities of the channel variances.

\begin{table}[t]
\centering
\vspace{-12pt}
\scalebox{0.95}{
  \begin{tabular}{@{}p{2.3cm}p{1.075cm}p{1.075cm}p{1.1cm}p{1.5cm}@{}}
    \toprule
    Method & \#Param.  & sMNIST & pMNIST & sCIFAR10 \\
    
    \midrule
    DilRNN~\cite{dilatedRNN2017}       & 44K    & 98.0    & 97.2    & -  \\
    LSTM~\cite{tcn2018}                & 70K    & 87.2    & 85.7    & -  \\
    GRU~\cite{tcn2018}                 & 70K    & 96.2    & 87.3    & -  \\
    TCN~\cite{tcn2018}                 & 70K    & 99.0    & 97.2    & -  \\
    FlexTCN-6~\cite{flexconv2021}      & 375K   & 99.6   & 98.6   & 80.8  \\
    r-LSTM~\cite{rLSTM2018}            & 500K   & 98.4    & 95.2    & 72.2  \\
    Transformer~\cite{rLSTM2018}       & 500K   & 98.9    & 97.9    & 62.2  \\
    HiPPO~\cite{hippo2020}             & 500K   & 98.9    & 98.3    & 61.1  \\
    CKCNN~\cite{ckconv2021}            & 1M     & 99.32   & 98.54   & 63.74  \\
    CCNN~\cite{ccnn2022}               & 2M     & \underline{99.72}   & \underline{98.84}   & \underline{93.08}  \\
    LSSL~\cite{lssl2021}               & 2M   & 99.53   & 98.76   & 84.65  \\  
    S4~\cite{S4-2021}                  & 7.8M   & 99.63   & 98.70   & 91.13  \\
    Hyena~\cite{hyena2023}             & 7.8M   & -   & -   & 91.10  \\
    TrellisNet~\cite{trellis2018}      & 8M     & 99.20   & 98.13   & 73.42  \\
    LRU~\cite{ssm-LRU2023}             & -   & -   & -   & 89.00  \\
    Liquid-S4~\cite{liquidS4-2022}  & -   & -   & -   & 92.02  \\
    S4D-Inv~\cite{S4D-2022}            & -   & -   & -   & 90.69  \\
    S5~\cite{S5-2022}                  & -   & 99.65   & 98.67   & 90.10  \\
    
    \midrule
    \multirow{1}*{\textbf{Terminator}} 
     & 1.3M   & \textbf{99.72}   & \textbf{99.12}   & \textbf{93.21}  \\
    \bottomrule
  \end{tabular}}
  \vspace{-3pt}
  \caption{Pixel-level 1D image classification. The underlined result is the second best method.}
  \label{tab:image-1d}
  \vspace{-11pt}
\end{table}

\subsection{Slow Neural Loss}
\label{sec:slow_neural_loss}
\vspace{-2pt}
The slow neural loss serves as a local feedback error for updating the slow network (i.e. large hyper-kernel generator). We compute the Mean Squared Error (MSE) of the $j$-th SFNE block as follows:
\begin{align}
\mathbb{L}_s^j &= \sum_{t=0}^{j-1} \| \mathbf{K}_g^{j} - \mathbb{E}(\mathbf{K}_g^{t})\|^2, j = 1, ..., J \\
\mathbb{L}_s &= \sum_{j=1}^{J} \mathbb{L}_s^j,
\end{align}
where $J$ is the number of SFNE blocks, and $\mathbb{E}$ represents a channel expansion operation achieved through repeated concatenations for different blocks ($t=0, ..., j-1$). It can exploit the channel-spatial consistency of $K_g$ in deep and shallow blocks to promote context-dependent $\hat{\mathbf{K}}_g$ to obtain more accurate pixel-level scores (see Figure~\ref{fig:vis_no_slow_loss}). The total slow neural loss $\mathbb{L}_s$ is obtained by summing up the individual block losses, $\mathbb{L}_s^j$ ($j=1, ..., J$).
In our experiments, we combine the $\mathbb{L}_s$ loss with the cross-entropy loss to train the model end-to-end using backpropagation.

\begin{table}[t]
\vspace{-12pt}
\centering
  \begin{tabular}{@{}lccc@{}}
    \toprule
    Method   & \#Param.  & CIFAR10 & CIFAR100 \\
    
    \midrule
    InceptionV2~\cite{inceptionV3-2016}      & 65M	 & -    & 72.49   \\  
    Xception~\cite{Xception2017}	         & 21M	 & -    & \underline{74.93}   \\
    VGG16~\cite{vgg2014}                     & 20M     & 93.91   & 74.08   \\
    ViT~\cite{ViT2020}                       & 6.3M   & 90.92   & 66.54   \\  
    ConvNeXt~\cite{convnext2022}             & 6.3M   & 92.20   & -   \\ 
    WRN-40-2~\cite{wide-resnet2016}          & 2.2M    & \underline{94.67}   & -    \\
    ResNet-110~\cite{resnet2016}             & 1.7M    & 93.57   & 74.84    \\
    
    \midrule
    Hyena-ISO~\cite{hyena2023}               & 202K   & 91.20   & -  \\
    CKCNN~\cite{ckconv2021}                  & 670K   & 86.80   & -   \\
    FlexNet-16~\cite{flexconv2021}           & 670K   & 92.20   & -  \\
    CCNN$^\dag$~\cite{ccnn2022}              & 2.0M   & 94.56   & 73.58   \\
    S4ND-ISO~\cite{s4nd2022}                 & 5.3M   & 94.10   & -   \\

    \midrule
    \multirow{1}*{\textbf{Terminator}} 
    & 1.3M   & \textbf{95.22}   & \textbf{75.38}   \\
    \bottomrule
  \end{tabular}
  \vspace{-3pt}
  \caption{2D image classification on CIFAR10 and CIFAR100 datasets. $^\dag$ represents the results obtained from the code released by~\cite{ccnn2022}.}
  \label{tab:image-2d}
  \vspace{-11pt}
\end{table}

\begin{table*}[t]
\vspace{1pt}
  \centering
  \begin{tabular}{@{}lcc|ccc@{}}
    \toprule

    Method      &  ResNet-152~\cite{resnet2016} \textit{w/} BN~\cite{BN2015}    &  ResNet-152~\cite{resnet2016} \textit{w/ BS}   & \textbf{Terminator}   & \textit{w/ Affine}   & \textit{w/ Momentum}    \\

    \hline
    Test Accuracy    & \textbf{85.41}    & 77.82    & \textbf{86.32}    & 85.93    & 85.55  \\

    \bottomrule
  \end{tabular}
  \vspace{-4pt}
  \caption{Comparison of test accuracy of ResNet-152 and Terminator on STL10 dataset when using normalization and standardization operations. ResNet-152 performs poorly when removing the affine and momentum parameters, while our Terminator does the opposite.}
  \label{tab:bs-is}
\vspace{-9pt}
\end{table*}

\begin{table}[t]
\vspace{5pt}
  \centering
  \begin{tabular}{@{}l|ccc|c@{}}
    \toprule

    Method   & Size  & \#Param.     & FLOPs  & Acc.   \\

    \hline
    FlexNet-16~\cite{flexconv2021}   & $96^2$   & 670K    & -   & 68.67 \\  
    CCNN$^\dag$~\cite{ccnn2022}     & $256^2$       & 2M       & 115G   & 78.01     \\  
    ResNet-18~\cite{EBN2020}   & $96^2$       & 11.2M      & 5G  & 78.65   \\

    DenseNet-201~\cite{densenet2017}    & $96^2$   & 18.1M      & 13G & 80.60   \\  

    Xception~\cite{Xception2017}     & $96^2$       & 20.8M     & 10G & 84.15    \\  

    ResNeXt-152~\cite{resnext2017}   & $96^2$  & 33.3M      & 18G  & 84.95    \\  

    ResNet-152~\cite{resnet2016}     & $96^2$  & 58.2M      & 34G  & \underline{85.41}    \\

    \hline
    \multirow{2}*{\textbf{Terminator}} 
    & $96^2$   & 1.5M       & 10G  & \textbf{80.13}   \\ 
    & $96^2$   & 7.8M       & 67G  & \textbf{86.32}   \\ 
    \bottomrule
  \end{tabular}
  \vspace{-4pt}
  \caption{Image classification on STL10 dataset. $^\dag$ represents the reproduced results obtained from the code released by~\cite{ccnn2022}.}
  \label{tab:image-stl}
\vspace{-12pt}
\end{table}

\section{Experiments}
\vspace{-2pt}
In this section, we provide quantitative and qualitative results to demonstrate the superior performance of our Terminator architecture. The experimental settings, dataset introduction, architectural details, ablation studies and more discussions are presented in the appendix.

\subsection{Results on 1D Image Classification}
\vspace{-2pt}
The results shown in Table~\ref{tab:image-1d} highlight the superior performance of our Terminator architecture on the sMNIST, pMNIST~\cite{mnist1998}, and sCIFAR10~\cite{cifar2009} datasets. Our network achieves state-of-the-art results, surpassing temporal convolutional networks~\cite{trellis2018}, continuous convolutional networks~\cite{flexconv2021,ckconv2021,ccnn2022} and state space models~\cite{hippo2020,lssl2021,S4-2021,liquidS4-2022,S4D-2022,S5-2022,hyena2023} in terms of accuracy, despite having fewer model parameters and not utilizing residual learning. 
One of the key factors contributing to our model's success is the $\mathrm{Hyper}\mathcal{Z{\cdot}Z{\cdot}W}$ operator, which effectively models long-term dependencies through context-dependent fast weights.

\subsection{Results on 2D Image Classification}
\vspace{-2pt}
Table~\ref{tab:image-2d} shows the performance comparison of our Terminator with various models on the CIFAR10 and CIFAR100 datasets. Remarkably, our Terminator model surpasses Inception models~\cite{inceptionV3-2016,Xception2017}, residual networks~\cite{resnet2016,wide-resnet2016,convnext2022}, and vision transformers~\cite{ViT2020} in terms of accuracy while utilizing fewer model parameters. Moreover, our network outperforms continuous convolutional networks~\cite{flexconv2021,ckconv2021,ccnn2022} and state space models~\cite{s4nd2022,hyena2023}, establishing state-of-the-art results without relying on residual learning.

Furthermore, the results in Table~\ref{tab:image-stl} demonstrate that our Terminator achieves superior performance compared to ResNet-152 while utilizing only $1/7$ of the model parameters. Notably, the Terminator model doubles the computation while efficiently maintaining full context interaction at each layer without information loss, even on large-resolution images. Consequently, the need for residual learning is eliminated. Additionally, our $\mathrm{Hyper}\mathcal{Z{\cdot}Z{\cdot}W}$ operator utilizes hidden activations to generate context-dependent fast weights, thereby enhancing the model's capacity for in-context learning.

\vspace{-3pt}
\subsection{Analysis of Excellent Properties}
\vspace{-3pt}

\textbf{Training Convergence.} The visualization in Figure~\ref{fig:stl-acc_bn_bs}(a) demonstrates the faster training convergence of our Terminator architecture. It achieves the same test accuracy as ResNet-152 in approximately $1/6$ of the epochs, requiring only $1/2$ of the training epochs. Moreover, it is noteworthy that our Terminator achieves satisfactory results on the sMNIST dataset with just 50 training epochs, which is only $1/4$ of the epochs required by other networks~\cite{ckconv2021,flexconv2021,ccnn2022,lssl2021,S4-2021}. The fast convergence of our model is mainly due to the zero-mean features introduced below, which is a well-known conclusion~\cite{lecun2002,BN2015}.

\textbf{Zero-mean Features.} The results presented in Figure~\ref{fig:stl-acc_bn_bs}(b) and Table~\ref{tab:bs-is} demonstrate that our network achieves stable zero-mean features and improved performance even without the $\textit{affine}$ and \textit{momentum} parameters in standardization layers. In contrast, training ResNet-152 with batch standardization (BS) leads to a significant decline in performance, resulting in an accuracy of only $77.82\%$. Additionally, the channel mean and channel variance experience notable increases. These elevated batch statistics introduce higher volatility, which detrimentally affects the model's convergence and generalization capabilities.

\textbf{Model Parameters.} The statistical findings in Table~\ref{tab:model_params} reveal that as the model depth increases, the number of model parameters in each block also increases, with the last block being particularly significant, accounting for 68\% of the total model parameters. This growth in parameters is primarily attributed to the escalating number of channels in the feature maps. Notably, the slow network has a mere 8\% of the model parameters. We also would like to highlight that, as demonstrated in Figure~\ref{fig:vis_res_termi}, our Terminator network no longer need to be very deep, because it can achieve effective image denoising by utilizing only a few SFNE blocks.


\section{Conclusion}
\vspace{-0.85pt}
This paper introduces the Terminator architecture, which offers a novel approach to network design by abandoning residual learning and leveraging large implicit convolution kernels. The SFNE block provides a new perspective for building multi-branch networks, while the $\mathrm{Hyper}\mathcal{Z{\cdot}Z{\cdot}W}$ operator enables dynamic encoding by providing context-dependent weights for the fast networks through elementwise multiplication between hyper-kernels and hidden activations. Our network brings several advantages, including faster training convergence, stable zero-mean features, and fewer model parameters. The experimental results demonstrate its superiority in capturing long-range dependencies and achieving state-of-the-art performance on pixel-level 1D and 2D image classification benchmarks. Due to limited computing resources, we were unable to conduct experiments on ImageNet dataset. However, in future research, we plan to prioritize the exploration of more effective slow neural loss to improve the accuracy of pixel-level scores.

\clearpage
{
    \small
    \bibliographystyle{ieeenat_fullname}
    \bibliography{main}

\begin{thebibliography}{51}
\providecommand{\natexlab}[1]{#1}
\providecommand{\url}[1]{\texttt{#1}}
\expandafter\ifx\csname urlstyle\endcsname\relax
  \providecommand{\doi}[1]{doi: #1}\else
  \providecommand{\doi}{doi: \begingroup \urlstyle{rm}\Url}\fi

\bibitem[Ba et~al.(2016{\natexlab{a}})Ba, Hinton, Mnih, Leibo, and Ionescu]{fwp-Ba2016}
Jimmy Ba, Geoffrey~E. Hinton, Volodymyr Mnih, Joel~Z. Leibo, and Catalin Ionescu.
\newblock Using fast weights to attend to the recent past.
\newblock In \emph{Neural Information Processing Systems}, 2016{\natexlab{a}}.

\bibitem[Ba et~al.(2016{\natexlab{b}})Ba, Kiros, and Hinton]{LN2016}
Jimmy~Lei Ba, Jamie~Ryan Kiros, and Geoffrey~E Hinton.
\newblock Layer normalization.
\newblock \emph{arXiv preprint arXiv:1607.06450}, 2016{\natexlab{b}}.

\bibitem[Bai et~al.(2018{\natexlab{a}})Bai, Kolter, and Koltun]{tcn2018}
Shaojie Bai, J~Zico Kolter, and Vladlen Koltun.
\newblock An empirical evaluation of generic convolutional and recurrent networks for sequence modeling.
\newblock \emph{arXiv preprint arXiv:1803.01271}, 2018{\natexlab{a}}.

\bibitem[Bai et~al.(2018{\natexlab{b}})Bai, Kolter, and Koltun]{trellis2018}
Shaojie Bai, J~Zico Kolter, and Vladlen Koltun.
\newblock Trellis networks for sequence modeling.
\newblock \emph{arXiv preprint arXiv:1810.06682}, 2018{\natexlab{b}}.

\bibitem[Chang et~al.(2017)Chang, Zhang, Han, Yu, Guo, Tan, Cui, Witbrock, Hasegawa-Johnson, and Huang]{dilatedRNN2017}
Shiyu Chang, Yang Zhang, Wei Han, Mo Yu, Xiaoxiao Guo, Wei Tan, Xiaodong Cui, Michael Witbrock, Mark~A Hasegawa-Johnson, and Thomas~S Huang.
\newblock Dilated recurrent neural networks.
\newblock \emph{Advances in neural information processing systems}, 30, 2017.

\bibitem[Chollet(2017)]{Xception2017}
Fran{\c{c}}ois Chollet.
\newblock Xception: Deep learning with depthwise separable convolutions.
\newblock In \emph{Proceedings of the IEEE conference on computer vision and pattern recognition}, pages 1251--1258, 2017.

\bibitem[Dauphin et~al.(2017)Dauphin, Fan, Auli, and Grangier]{glu2017}
Yann~N Dauphin, Angela Fan, Michael Auli, and David Grangier.
\newblock Language modeling with gated convolutional networks.
\newblock In \emph{International conference on machine learning}, pages 933--941. PMLR, 2017.

\bibitem[Dosovitskiy et~al.(2020)Dosovitskiy, Beyer, Kolesnikov, Weissenborn, Zhai, Unterthiner, Dehghani, Minderer, Heigold, Gelly, et~al.]{ViT2020}
Alexey Dosovitskiy, Lucas Beyer, Alexander Kolesnikov, Dirk Weissenborn, Xiaohua Zhai, Thomas Unterthiner, Mostafa Dehghani, Matthias Minderer, Georg Heigold, Sylvain Gelly, et~al.
\newblock An image is worth 16x16 words: Transformers for image recognition at scale.
\newblock \emph{arXiv preprint arXiv:2010.11929}, 2020.

\bibitem[Fathony et~al.(2020)Fathony, Sahu, Willmott, and Kolter]{mfn2020}
Rizal Fathony, Anit~Kumar Sahu, Devin Willmott, and J~Zico Kolter.
\newblock Multiplicative filter networks.
\newblock In \emph{International Conference on Learning Representations}, 2020.

\bibitem[Gu et~al.(2020)Gu, Dao, Ermon, Rudra, and R{\'e}]{hippo2020}
Albert Gu, Tri Dao, Stefano Ermon, Atri Rudra, and Christopher R{\'e}.
\newblock Hippo: Recurrent memory with optimal polynomial projections.
\newblock \emph{Advances in neural information processing systems}, 33:\penalty0 1474--1487, 2020.

\bibitem[Gu et~al.(2021{\natexlab{a}})Gu, Goel, and R{\'e}]{S4-2021}
Albert Gu, Karan Goel, and Christopher R{\'e}.
\newblock Efficiently modeling long sequences with structured state spaces.
\newblock \emph{arXiv preprint arXiv:2111.00396}, 2021{\natexlab{a}}.

\bibitem[Gu et~al.(2021{\natexlab{b}})Gu, Johnson, Goel, Saab, Dao, Rudra, and R{\'e}]{lssl2021}
Albert Gu, Isys Johnson, Karan Goel, Khaled Saab, Tri Dao, Atri Rudra, and Christopher R{\'e}.
\newblock Combining recurrent, convolutional, and continuous-time models with linear state space layers.
\newblock \emph{Advances in neural information processing systems}, 34:\penalty0 572--585, 2021{\natexlab{b}}.

\bibitem[Gu et~al.(2022)Gu, Goel, Gupta, and R{\'e}]{S4D-2022}
Albert Gu, Karan Goel, Ankit Gupta, and Christopher R{\'e}.
\newblock On the parameterization and initialization of diagonal state space models.
\newblock \emph{Advances in Neural Information Processing Systems}, 35:\penalty0 35971--35983, 2022.

\bibitem[Hasani et~al.(2022)Hasani, Lechner, Wang, Chahine, Amini, and Rus]{liquidS4-2022}
Ramin Hasani, Mathias Lechner, Tsun-Hsuan Wang, Makram Chahine, Alexander Amini, and Daniela Rus.
\newblock Liquid structural state-space models.
\newblock \emph{arXiv preprint arXiv:2209.12951}, 2022.

\bibitem[He et~al.(2016)He, Zhang, Ren, and Sun]{resnet2016}
Kaiming He, Xiangyu Zhang, Shaoqing Ren, and Jian Sun.
\newblock Deep residual learning for image recognition.
\newblock In \emph{Proceedings of the IEEE conference on computer vision and pattern recognition}, pages 770--778, 2016.

\bibitem[Hinton and Plaut(1987)]{slow-fast1987}
Geoffrey~E Hinton and David~C Plaut.
\newblock Using fast weights to deblur old memories.
\newblock In \emph{Proceedings of the ninth annual conference of the Cognitive Science Society}, pages 177--186, 1987.

\bibitem[Hu et~al.(2018)Hu, Shen, and Sun]{senet2018}
Jie Hu, Li Shen, and Gang Sun.
\newblock Squeeze-and-excitation networks.
\newblock In \emph{Proceedings of the IEEE conference on computer vision and pattern recognition}, pages 7132--7141, 2018.

\bibitem[Huang et~al.(2017)Huang, Liu, Van Der~Maaten, and Weinberger]{densenet2017}
Gao Huang, Zhuang Liu, Laurens Van Der~Maaten, and Kilian~Q Weinberger.
\newblock Densely connected convolutional networks.
\newblock In \emph{Proceedings of the IEEE conference on computer vision and pattern recognition}, pages 4700--4708, 2017.

\bibitem[Ioffe and Szegedy(2015)]{BN2015}
Sergey Ioffe and Christian Szegedy.
\newblock Batch normalization: Accelerating deep network training by reducing internal covariate shift.
\newblock In \emph{International conference on machine learning}, pages 448--456. pmlr, 2015.

\bibitem[Krizhevsky et~al.(2009)Krizhevsky, Hinton, et~al.]{cifar2009}
Alex Krizhevsky, Geoffrey Hinton, et~al.
\newblock Learning multiple layers of features from tiny images.
\newblock 2009.

\bibitem[LeCun(1998)]{mnist1998}
Yann LeCun.
\newblock The mnist database of handwritten digits.
\newblock \emph{http://yann. lecun. com/exdb/mnist/}, 1998.

\bibitem[LeCun et~al.(2002)LeCun, Bottou, Orr, and M{\"u}ller]{lecun2002}
Yann LeCun, L{\'e}on Bottou, Genevieve~B Orr, and Klaus-Robert M{\"u}ller.
\newblock Efficient backprop.
\newblock In \emph{Neural networks: Tricks of the trade}, pages 9--50. Springer, 2002.

\bibitem[Liu et~al.(2021)Liu, Lin, Cao, Hu, Wei, Zhang, Lin, and Guo]{swin2021}
Ze Liu, Yutong Lin, Yue Cao, Han Hu, Yixuan Wei, Zheng Zhang, Stephen Lin, and Baining Guo.
\newblock Swin transformer: Hierarchical vision transformer using shifted windows.
\newblock In \emph{Proceedings of the IEEE/CVF international conference on computer vision}, pages 10012--10022, 2021.

\bibitem[Liu et~al.(2022)Liu, Mao, Wu, Feichtenhofer, Darrell, and Xie]{convnext2022}
Zhuang Liu, Hanzi Mao, Chao-Yuan Wu, Christoph Feichtenhofer, Trevor Darrell, and Saining Xie.
\newblock A convnet for the 2020s.
\newblock In \emph{Proceedings of the IEEE/CVF conference on computer vision and pattern recognition}, pages 11976--11986, 2022.

\bibitem[Luo et~al.(2020)Luo, Zhan, Wang, and Gao]{EBN2020}
Chunjie Luo, Jianfeng Zhan, Lei Wang, and Wanling Gao.
\newblock Extended batch normalization.
\newblock \emph{arXiv preprint arXiv:2003.05569}, 2020.

\bibitem[Nguyen et~al.(2022)Nguyen, Goel, Gu, Downs, Shah, Dao, Baccus, and R{\'e}]{s4nd2022}
Eric Nguyen, Karan Goel, Albert Gu, Gordon Downs, Preey Shah, Tri Dao, Stephen Baccus, and Christopher R{\'e}.
\newblock S4nd: Modeling images and videos as multidimensional signals with state spaces.
\newblock \emph{Advances in neural information processing systems}, 35:\penalty0 2846--2861, 2022.

\bibitem[N{\o}kland and Eidnes(2019)]{local2019}
Arild N{\o}kland and Lars~Hiller Eidnes.
\newblock Training neural networks with local error signals.
\newblock In \emph{International conference on machine learning}, pages 4839--4850. PMLR, 2019.

\bibitem[Orvieto et~al.(2023)Orvieto, Smith, Gu, Fernando, Gulcehre, Pascanu, and De]{ssm-LRU2023}
Antonio Orvieto, Samuel~L Smith, Albert Gu, Anushan Fernando, Caglar Gulcehre, Razvan Pascanu, and Soham De.
\newblock Resurrecting recurrent neural networks for long sequences.
\newblock \emph{arXiv preprint arXiv:2303.06349}, 2023.

\bibitem[Poli et~al.(2023)Poli, Massaroli, Nguyen, Fu, Dao, Baccus, Bengio, Ermon, and R{\'e}]{hyena2023}
Michael Poli, Stefano Massaroli, Eric Nguyen, Daniel~Y Fu, Tri Dao, Stephen Baccus, Yoshua Bengio, Stefano Ermon, and Christopher R{\'e}.
\newblock Hyena hierarchy: Towards larger convolutional language models.
\newblock \emph{arXiv preprint arXiv:2302.10866}, 2023.

\bibitem[Romero et~al.(2021{\natexlab{a}})Romero, Bruintjes, Tomczak, Bekkers, Hoogendoorn, and van Gemert]{flexconv2021}
David~W Romero, Robert-Jan Bruintjes, Jakub~M Tomczak, Erik~J Bekkers, Mark Hoogendoorn, and Jan~C van Gemert.
\newblock Flexconv: Continuous kernel convolutions with differentiable kernel sizes.
\newblock \emph{arXiv preprint arXiv:2110.08059}, 2021{\natexlab{a}}.

\bibitem[Romero et~al.(2021{\natexlab{b}})Romero, Kuzina, Bekkers, Tomczak, and Hoogendoorn]{ckconv2021}
David~W Romero, Anna Kuzina, Erik~J Bekkers, Jakub~M Tomczak, and Mark Hoogendoorn.
\newblock Ckconv: Continuous kernel convolution for sequential data.
\newblock \emph{arXiv preprint arXiv:2102.02611}, 2021{\natexlab{b}}.

\bibitem[Romero et~al.(2022)Romero, Knigge, Gu, Bekkers, Gavves, Tomczak, and Hoogendoorn]{ccnn2022}
David~W Romero, David~M Knigge, Albert Gu, Erik~J Bekkers, Efstratios Gavves, Jakub~M Tomczak, and Mark Hoogendoorn.
\newblock Towards a general purpose cnn for long range dependencies in $ n $ d.
\newblock 2022.

\bibitem[Rudin et~al.(2022)Rudin, Chen, Chen, Huang, Semenova, and Zhong]{ex-challenge2022}
Cynthia Rudin, Chaofan Chen, Zhi Chen, Haiyang Huang, Lesia Semenova, and Chudi Zhong.
\newblock Interpretable machine learning: Fundamental principles and 10 grand challenges.
\newblock \emph{Statistic Surveys}, 16:\penalty0 1--85, 2022.

\bibitem[Salimans and Kingma(2016)]{weight-norm2016}
Tim Salimans and Durk~P Kingma.
\newblock Weight normalization: A simple reparameterization to accelerate training of deep neural networks.
\newblock \emph{Advances in neural information processing systems}, 29, 2016.

\bibitem[Schlag et~al.(2021)Schlag, Irie, and Schmidhuber]{fwp2021}
Imanol Schlag, Kazuki Irie, and J{\"u}rgen Schmidhuber.
\newblock Linear transformers are secretly fast weight programmers.
\newblock In \emph{International Conference on Machine Learning}, pages 9355--9366. PMLR, 2021.

\bibitem[Schmidhuber(1992)]{fwp-jurgen1992}
J{\"u}rgen Schmidhuber.
\newblock Learning to control fast-weight memories: An alternative to dynamic recurrent networks.
\newblock \emph{Neural Computation}, 4\penalty0 (1):\penalty0 131--139, 1992.

\bibitem[Simonyan and Zisserman(2014)]{vgg2014}
Karen Simonyan and Andrew Zisserman.
\newblock Very deep convolutional networks for large-scale image recognition.
\newblock \emph{arXiv preprint arXiv:1409.1556}, 2014.

\bibitem[Smith et~al.(2022)Smith, Warrington, and Linderman]{S5-2022}
Jimmy~TH Smith, Andrew Warrington, and Scott~W Linderman.
\newblock Simplified state space layers for sequence modeling.
\newblock \emph{arXiv preprint arXiv:2208.04933}, 2022.

\bibitem[Stanley et~al.(2009)Stanley, D'Ambrosio, and Gauci]{HyperNEAT2009}
Kenneth~O Stanley, David~B D'Ambrosio, and Jason Gauci.
\newblock A hypercube-based encoding for evolving large-scale neural networks.
\newblock \emph{Artificial life}, 15\penalty0 (2):\penalty0 185--212, 2009.

\bibitem[Szegedy et~al.(2016)Szegedy, Vanhoucke, Ioffe, Shlens, and Wojna]{inceptionV3-2016}
Christian Szegedy, Vincent Vanhoucke, Sergey Ioffe, Jon Shlens, and Zbigniew Wojna.
\newblock Rethinking the inception architecture for computer vision.
\newblock In \emph{Proceedings of the IEEE conference on computer vision and pattern recognition}, pages 2818--2826, 2016.

\bibitem[Tan and Le(2019)]{efficientnet2019}
Mingxing Tan and Quoc Le.
\newblock Efficientnet: Rethinking model scaling for convolutional neural networks.
\newblock In \emph{International conference on machine learning}, pages 6105--6114. PMLR, 2019.

\bibitem[Trinh et~al.(2018)Trinh, Dai, Luong, and Le]{rLSTM2018}
Trieu Trinh, Andrew Dai, Thang Luong, and Quoc Le.
\newblock Learning longer-term dependencies in rnns with auxiliary losses.
\newblock In \emph{International Conference on Machine Learning}, pages 4965--4974. PMLR, 2018.

\bibitem[Ulyanov et~al.(2016)Ulyanov, Vedaldi, and Lempitsky]{IN2016}
Dmitry Ulyanov, Andrea Vedaldi, and Victor Lempitsky.
\newblock Instance normalization: The missing ingredient for fast stylization.
\newblock \emph{arXiv preprint arXiv:1607.08022}, 2016.

\bibitem[Vaswani et~al.(2017)Vaswani, Shazeer, Parmar, Uszkoreit, Jones, Gomez, Kaiser, and Polosukhin]{attention2017}
Ashish Vaswani, Noam~M. Shazeer, Niki Parmar, Jakob Uszkoreit, Llion Jones, Aidan~N. Gomez, Lukasz Kaiser, and Illia Polosukhin.
\newblock Attention is all you need.
\newblock In \emph{Neural Information Processing Systems}, 2017.

\bibitem[Veit et~al.(2016)Veit, Wilber, and Belongie]{ex-residual-ensemble2016}
Andreas Veit, Michael~J Wilber, and Serge Belongie.
\newblock Residual networks behave like ensembles of relatively shallow networks.
\newblock \emph{Advances in neural information processing systems}, 29, 2016.

\bibitem[Wang et~al.(2020)Wang, Wu, Zhu, Li, Zuo, and Hu]{eca2020}
Qilong Wang, Banggu Wu, Pengfei Zhu, Peihua Li, Wangmeng Zuo, and Qinghua Hu.
\newblock Eca-net: Efficient channel attention for deep convolutional neural networks.
\newblock In \emph{Proceedings of the IEEE/CVF conference on computer vision and pattern recognition}, pages 11534--11542, 2020.

\bibitem[Wu and He(2018)]{GN2018}
Yuxin Wu and Kaiming He.
\newblock Group normalization.
\newblock In \emph{Proceedings of the European conference on computer vision (ECCV)}, pages 3--19, 2018.

\bibitem[Xie et~al.(2017)Xie, Girshick, Doll{\'a}r, Tu, and He]{resnext2017}
Saining Xie, Ross Girshick, Piotr Doll{\'a}r, Zhuowen Tu, and Kaiming He.
\newblock Aggregated residual transformations for deep neural networks.
\newblock In \emph{Proceedings of the IEEE conference on computer vision and pattern recognition}, pages 1492--1500, 2017.

\bibitem[Zagoruyko and Komodakis(2016)]{wide-resnet2016}
Sergey Zagoruyko and Nikos Komodakis.
\newblock Wide residual networks.
\newblock \emph{arXiv preprint arXiv:1605.07146}, 2016.

\bibitem[Zagoruyko and Komodakis(2017)]{diracnets2017}
Sergey Zagoruyko and Nikos Komodakis.
\newblock Diracnets: Training very deep neural networks without skip-connections.
\newblock \emph{arXiv preprint arXiv:1706.00388}, 2017.

\bibitem[Zhai et~al.(2021)Zhai, Talbott, Srivastava, Huang, Goh, Zhang, and Susskind]{attention-free2021}
Shuangfei Zhai, Walter Talbott, Nitish Srivastava, Chen Huang, Hanlin Goh, Ruixiang Zhang, and Josh Susskind.
\newblock An attention free transformer.
\newblock \emph{arXiv preprint arXiv:2105.14103}, 2021.

\end{thebibliography}
}





\end{document}